\documentclass[letterpaper, 10 pt, conference]{ieeeconf}  

\usepackage{url,graphicx,tabularx,array,verbatim,multirow,epstopdf,amsmath,amssymb,mathrsfs,bm,hyperref}
\usepackage[tableposition=top]{caption}
\usepackage{float}
\floatstyle{plaintop}
\restylefloat{table}
\usepackage{subcaption}
\usepackage{tikz}
\usepackage{pgfplots}
\newcommand{\subparagraph}{}
\usetikzlibrary{spy,calc}
\usepackage{titlesec}

\IEEEoverridecommandlockouts                              
\renewcommand{\baselinestretch}{0.99}
\titlespacing*{\section}{0pt}{0.5\baselineskip}{0.5\baselineskip}
\titlespacing*{\subsection}{0pt}{0.5\baselineskip}{0.5\baselineskip}

\title{\LARGE \bf
State Estimation for a Humanoid Robot}

\author{Nicholas Rotella$^{1}$, Michael Bloesch$^{2}$, Ludovic Righetti$^{3}$ and Stefan Schaal$^{1,3}$
\thanks{This research was supported in part by National Science Foundation grants ECS-0326095, IIS-0535282, IIS-1017134, CNS-0619937, IIS-0917318, CBET-0922784, EECS-0926052, CNS-0960061, the DARPA program on Autonomous Robotic Manipulation, the Army Research Office, the Okawa Foundation, the ATR Computational Neuroscience Laboratories, the Max-Planck-Society, the Swiss National Science Foundation (SNF) through project 200021\_149427 / 1 and the National Centre of Competence in Research Robotics.}
\thanks{$^{1}$Computational Learning and Motor Control Lab, University of Southern California, Los Angeles, California.}
\thanks{$^{2}$Autonomous Systems Lab, ETH Zurich, Zurich, Switzerland.}
\thanks{$^{3}$Autonomous Motion Department, Max Planck Institute for Intelligent Systems, Tuebingen, Germany.}
}

\begin{document}

\maketitle
\thispagestyle{empty}
\pagestyle{empty}

\begin{abstract}

This paper introduces a framework for state estimation on a humanoid robot
platform using only common proprioceptive sensors and knowledge of
leg kinematics.  The presented approach extends that detailed in prior work on a
point-foot quadruped platform by adding the rotational constraints imposed by the
humanoid's flat feet.  As in previous work, the proposed Extended Kalman Filter accommodates contact switching and makes no assumptions about gait
or terrain, making it applicable on any humanoid platform for use in any
task.  A nonlinear
observability analysis is performed on both the point-foot and flat-foot filters
and it is concluded that the addition of rotational constraints significantly simplifies singular
cases and improves the observability characteristics of the system.  Results on
a simulated walking dataset demonstrate the performance gain of the
flat-foot filter as well as confirm the results of the presented observability
analysis.
\end{abstract}


\section{INTRODUCTION}

State estimation has long been a topic of importance in mobile
robotics, where the typical filter architecture fuses wheel odometry (also known as ``dead
reckoning'') with external sensors in order to correct for errors due to
wheel slippage and other factors.  In most cases, state estimation entails maintaining an estimate of the robot's absolute position and yaw for navigation in simple environments.  

Unlike traditional mobile robot platforms, however, legged robots usually
require knowledge of the full 6DOF pose of the base for control.  Further, the utility of such platforms is their potential for operation in unstructured environments.  Exteroceptive sensors such as cameras
or GPS  units are unfit for use in such situations, forcing the tasks of
control and localization to depend on internal (proprioceptive) sensing.  While
wheeled robots are assumed to remain stable and in contact at all times, legged robot locomotion inherently involves
intermittent contacts.  This makes stability a main concern as well as complicates odometry-based estimation approaches. The problem of state estimation for legged robots is thus fundamentally different than that of estimation for wheeled robots.
The goal of this work is thus to build on \cite{main} to develop a state estimation framework suitable for the task of bipedal locomotion.  As such, we review traditional approaches in order to better define the goals of this work.

At its simplest, state estimation entails determination of the transformation describing the robot's global pose.  One of the earliest attempts \cite{krotkov} was performed on the CMU Ambler, a hexapod robot having only joint encoders.  The positions of the feet were computed both from motor commands and encoder measurements.  Each foot contributed a measurement and the transformation which minimized the error between the two estimates was computed.

Nearly fifteen years later, the approach detailed by Gassmann et al. in \cite{gassman} relied on the
same dead-reckoning method, extending earlier work by fusing odometry with orientation and position
measurements from new inertial sensors such as IMU and GPS
units.  However, this approach was inherently unable to handle
non statically-stable gaits. 

Around the same time, Lin et al. \cite{lin} extended previous work on pose estimation using a strain-based model with MEMS inertial sensors to measure motion of the body.  The gait was split into multiple phases, each of which corresponded to a simple Kalman filter. Models for each phase were switched using sensory cues, allowing non statically-stable gaits.  However, it was shown that the filter provides accurate pose estimates only when in tripod support.
 
Chitta et al. employed a particle filter which used an odometry-based prediction model
for the COM during quadruple support and an update model based on IMU data,
joint encoder readings and knowledge of the terrain relief for state estimation
with the quadruped LittleDog in \cite{chitta}.  While this method permitted global localization, it assumed knowledge of the terrain and a statically-stable gait.

Cobano et al. introduced a state estimator in \cite{cobano} for the quadruped SILO4 which
fused changes in position (computed using dead-reckoning) with magnetometer and
DGPS measurements in an Extended Kalman Filter (EKF).  While they tracked global position and heading without gait assumptions, the full 6DOF pose was not estimated.

In \cite{reinstein}, Reinstein and Hoffmann introduced an EKF for their
quadruped which combined kinematic predictions from IMU data with a ``data-driven'' velocity
measurement. By assuming an uninterrupted gait, the velocity was computed using a learned model.  However, the gait assumption and model limited their approach's utility on other platforms.

Recent work \cite{chilian} on the DLR Crawler hexapod platform by Chilian et al. introduced
an information filter suitable for combining multiple types of measurements.  The process model integrated IMU data to track the pose of the hexapod while visual and
leg odometry were used for updates.  Absolute measurements of roll and pitch angles were obtained from
the accelerometer and used as well. While they made no gait assumptions, their leg odometry measurements were valid only during periods of three or more contacts.

Biped state estimation has only recently gained attention.  Park et al. \cite{park} introduced a Kalman Filter in the context of a Zero Moment Point (ZMP)
balance controller using the Linear Inverted Pendulum Model (LIPM) to
approximate the dynamics of the humanoid robot.
The position, velocity, and acceleration of the COM of the LIPM were estimated using the pendulum dynamics and the measured ZMP location was used for the update step. However, their approach was only applicable in the context of ZMP balancing and walking.

In a similar context, Wang et al. proposed in \cite{wang} an Unscented Kalman Filter which
provided estimates of the joint angles and velocities for predictive ZMP control.  The filter treated the biped in single support as a fixed-base manipulator with corresponding dynamics.  Of course, this assumption was violated if the robot lost contact or slipped.  Additionally, the absolute orientation could not be observed.  This filter was also computationally-demanding as it used the full manipulator dynamics for prediction.

Given the above assessment of previous work, it is conjectured that a suitable
filter for general biped state estimation should 1) use only proprioceptive
sensors 2) make no assumptions about the gait or terrain 3) be
easily adapted to any humanoid platform and 4) use as little computational
resources as possible.  In \cite{main}, Bloesch et al. introduced a novel EKF for state estimation on the quadruped starlETH which fused leg odometry
and IMU data to estimate the full pose of the robot without making these assumptions.  Further, it was shown that as long as at least one foot is in contact with the ground then all
states other than absolute position and yaw (neither of which matters for stability) are observable.  Combined with
the fact that contact switching is easily handled, this result makes the filter applicable to bipeds which experience intermittent contacts and even transient aerial phases.  Because it assumes nothing about gait or terrain, the presented approach can be used on any legged robot in the context of any locomotion task.

The goal of this work is to adapt this approach to a humanoid platform by incorporating into the filter the rotational constraints provided by the flat feet of the biped.  Intuitively, a single flat foot contact fully constrains the pose of the robot - this suggests superior performance of the augmented filter in the
single support phase which is crucial for walking.  This extension is shown through theoretical analysis to improve the
observability properties of the filter and to increase the accuracy of the
estimation as demonstrated on simulated walking data having realistic noise
levels.

\section{Estimation Framework}

In order to introduce required terminology and notation, we begin by briefly
reviewing the problem of state estimation.  The Kalman Filter provides an estimate of the state vector $x$ along with a corresponding covariance matrix $P$ which specifies the uncertainty of the estimate.  The filter involves 1) propagating the estimates through the system in the \emph{prediction} step to produce \emph{a priori} estimates
$\hat{x}_{k}^{-}$ and $P_{k}^{-}$ and 2) updating the estimates using a measurement in the \emph{update step} to the \emph{a posteriori} estimates $\hat{x}_{k}^{+}$ and $P_{k}^{+}$.

However, the standard Kalman Filter is applicable only for state
estimation in linear systems.  Suppose we have the continuous-time, nonlinear
system
\begin{align}
\dot{x} &= f(x,w)\label{eq:nonlin_pred}\\
y &= h(x,v)\label{eq:nonlin_meas}
\end{align}
where $f()$ is the \emph{prediction} model$, h()$ is the \emph{measurement} model and $w$ and $v$ are noise terms.  One may simply linearize these models
around the current estimate and apply the Kalman Filter equations.  This is known as the \emph{Extended Kalman Filter}.  While optimality and
convergence are no longer guaranteed, this is a common approach for nonlinear estimation.  We chose the EKF over alternatives such as the Particle Filter or Unscented Kalman Filter for its simplicity and low computational cost.  However, the presented framework and observability analysis hold for all these filters.

\section{Handling of Rotational Quantities}

The unit quaternion was chosen to represent the base orientation in the original
filter due to its theoretical and computational advantages.  However, since the
quaternion is a non-minimal representation of $SO(3)$, special care must be taken
in handling rotational quantities in the EKF.

First, the exponential map
\begin{equation}
\exp{(\omega)} = 
\begin{pmatrix}
\sin{(\frac{||\omega||}{2}})\frac{\omega}{||\omega||}\\
\cos{(\frac{||\omega||}{2}})
\end{pmatrix}
\label{eq:exp}
\end{equation}
is used to relate a quaternion at times $k$ and $k+1$ given an incremental
rotation of magnitude $||\omega||$ about the unit vector $\omega/||\omega||$. 
That is, 
\begin{equation}
q_{k+1} = \exp{(\omega)}\otimes q_{k}
\end{equation}
where $\otimes$ denotes quaternion multiplication.  Roughly speaking, the
exponential map is used for ``addition'' of rotational quantities.  Note that the first entry of \eqref{eq:exp} is the vector portion of the quaternion while the second entry is the scalar portion.  Also note that there are two self-consistent conventions for quaternion algebra; to avoid such issues, we employ the quaternion conventions of \cite{trawny}.

In the EKF state vector, a quaternion is represented by its corresponding
three-dimensional \emph{error rotation vector} $\phi \in so(3)$.  The covariance of the
orientation represented by the quaternion is thus defined with respect to this
minimal representation. 

During the update step, the innovations vector $e$ corresponding to a
quaternion-valued measurement is computed as the three dimensional
rotation vector extracted from the difference between the actual measurement
quaternion $s$ and the expected measurement quaternion $z$.
That is,
\begin{equation}
e = \log{(s \otimes z^{-1})}
\label{eq:log}
\end{equation}
where $\log{()}$ denotes the logarithm mapping an element of $SO(3)$ to its
corresponding element of $so(3)$ (the inverse of the exponential map).  The above operation extends the notion of
subtraction to rotational quantities.

Finally, using the innovations vector as computed above, the state correction
vector $\Delta x$ is computed during the update step as $\Delta x = Ke$ where $K$ is the Kalman gain.  While
all non-rotational states are updated simply as $\hat{x}_{k}^{+} = \hat{x}_{k}^{-} +
\Delta x$, a quaternion state is updated using \eqref{eq:exp} as follows.
\begin{equation}
\hat{q}_{k}^{+} = \exp{(\Delta \phi)}\otimes \hat{q}_{k}^{-}
\end{equation}
For more information on quaternions and their use in state estimation see \cite{trawny}.

\section{Biped Prediction and Measurement Models}

In \cite{main} a continuous-time, nonlinear prediction model describing the time
evolution of the state of the quadruped was developed based on rigid body kinematics and a simple
model of an IMU consisting of a three-axis accelerometer and a three-axis
gyroscope.  This prediction model is shown below.  Note that this formulation
can accommodate an arbitrary number of feet, denoted $N$.
\begin{align}
\dot{r} &= v\\
\dot{v} &= a = C^{T}(\tilde{f} - b_{f} - w_{f}) + g\\
\dot{q} &= \frac{1}{2}\begin{bmatrix}\tilde{\omega} - b_{\omega} - w_{\omega}\\
0\end{bmatrix}\otimes q\\
 \dot{p}_{i} &= C^{T}w_{p,i} \quad \forall i \in \{1,\ldots,N\}
\\
\dot{b}_{f} &= w_{bf}\\
\dot{b}_{\omega} &= w_{b\omega}
\end{align}
The state of the filter is $x = [r, v, q, p_{i}, b_{f}, b_{w}]$ where $r$
is the position of the IMU (assumed to be located at the base), $v$
is the base velocity, $q$ is the quaternion representing a rotation from the world to body frame, $p_{i}$ is the world position of the $i^{th}$ foot and $b_{f}$ and $b_{w}$ are the accelerometer and gyroscope biases,
respectively.  Unless noted, $C$ represents the rotation
matrix corresponding to $q$.  The raw accelerometer and
gyroscope data are $\tilde{f}$ and $\tilde{\omega}$ and are
modeled as being subject to additive thermal noise processes $w_{f}$
and $w_{\omega}$ as well as random-walk biases parameterized by the noise
processes $w_{bf}$ and $w_{b\omega}$ (this is a common simple model in the inertial navigation literature which captures the main noise sources of IMU sensors).  Finally, $w_{p,i}$ denotes the noise
process representing the uncertainty of the $i^{th}$ foothold position.

To extend the established filter to a humanoid platform, we now augment the state vector with the orientations of the feet.
Including only one foot for brevity, the new state is defined as
$x = [r, v, q, p, b_{f}, b_{w}, z]$ where $z$ is the quaternion representing the
rotation from the world to foot frame.  Analogous to the assumption
made in \cite{main}, we assume that the orientation of
the foot remains constant while in contact; to permit a small amount of
rotational slippage, the prediction equation is defined to be 
\begin{equation}
\dot{z} = \frac{1}{2}\begin{bmatrix}w_{z}\\ 0\end{bmatrix}\otimes z
\end{equation}
where $w_{z}$ is the process noise having covariance matrix $Q_{z}$.  The
foot orientation noise is applied as an angular velocity in order to remain consistent with the original model.

The original filter update step was performed using one measurement equation for
each foot which represented the position of that foot relative to the base as
measured in the base frame.  This measurement is a function only of the
measured joint angles and the kinematic model of the leg; it is written as
\begin{equation}
s_{p} = C(p - r) + n_{p}\label{eq:pos_meas}
\end{equation}
The noise vector $n_{p}$ represents the combination of noise in the
encoders and uncertainty in the kinematic model.  Its covariance is the main
tuning parameter of the filter.

Following the original filter formulation, we introduce an additional
measurement which is again a function of only the measured joint angles and
the kinematics model.  The orientation of the foot in the base frame represented
by the quaternion
\begin{equation}
s_{z} = \exp{(n_{q})}\otimes q \otimes z^{-1}\label{eq:rot_meas}
\end{equation}
describes the rotational constraint imposed by a flat foot contact.  The noise
term $n_{q}$ is applied using the exponential map as in \eqref{eq:exp}.  Again, the
noise term depends on the forward kinematics uncertainty and constitutes a tuning parameter.

When a foot loses contact, its measurement equations are temporarily dropped from the filter. Additionally, its pose is removed from the state; this can be achieved more simply by setting the variances of $w_{p}$ and $w_{z}$ corresponding to the foot to large values. This causes the pose uncertainty to grow rapidly. When contact is restored, the pose and measurements are included again; this triggers a reset of the foot pose to its new value.  This allows for contact switching without the need for separate models.  During an aerial 
phase, the filter reduces to
integration of the prediction model.

\section{Implementation Details}

Since the above system is continuous and nonlinear, it must be discretized
and linearized for implementation purposes.  Following the EKF framework
outlined above, this requires two systems of equations: a \emph{discrete, nonlinear} system for prediction of
the state and measurement and a \emph{discrete, linear} system for propagation
of the state covariance through the prediction model and for computation of
the gain in the update step.
Additionally, we will derive in an intermediate step a continuous, linear
system.  We begin with a discussion of the discrete, nonlinear system used for prediction.

\subsection{Discrete, Nonlinear Model}

The first step in the EKF is the propagation of the expected value of the state
using the discretized nonlinear model.  Assuming a
zero-order hold on the IMU data over a small timestep $\Delta t$, we may discretize the original system using a first-order
integration scheme as
\begin{align}
\hat{r}_{k+1}^{-} &= \hat{r}_{k}^{+} + \Delta t \hat{v}_{k}^{+} + \frac{\Delta
t^{2}}{2}(\hat{C}_{k}^{+T}\hat{f}_{k}+g)\\
\hat{v}_{k+1}^{-} &= \hat{v}_{k}^{+} + \Delta t(\hat{C}_{k}^{+T}\hat{f}_{k}+g)\\
\hat{q}_{k+1}^{-} &= \exp{(\Delta t\hat{\omega}_{k})}\otimes \hat{q}_{k}^{+}\\
\hat{p}_{k+1}^{-} &= \hat{p}_{i,k}^{+}\\
\hat{b}_{f,k+1}^{-} &= \hat{b}_{f,k}^{+}\\
\hat{b}_{\omega,k+1}^{-} &= \hat{b}_{\omega,k}^{+}\\
\hat{z}_{k+1}^{-} &= \hat{z}_{i,k}^{+}
\end{align}

where $\hat{f}_{k} = \tilde{f}-\hat{b}_{f,k}^{+}$ and $\hat{\omega}_{k} =
\tilde{\omega}-\hat{b}_{\omega,k}^{+}$ denote the expected values of the measured
acceleration and angular velocity, respectively.
Note that the quaternion representing the base pose is updated using the exponential map formed from the infinitesimal rotation $\Delta t\hat{\omega}$ which is measured in the base frame directly.  Also note that a second-order discretization is used for the position in order to incorporate the
IMU acceleration at the current timestep.

The measurement model is discretized simply as
\begin{align}
\hat{s}_{p,k} &= \hat{C}_{k}^{-}(\hat{p}_{k}^{-} - \hat{r}_{k}^{-})\\
\hat{s}_{z,k} &= \hat{q}_{k}^{-} \otimes {\left(\hat{z}_{k}^{-}\right)}^{-1}
\end{align}
to produce the expected measurement of the nonlinear system.

\subsection{Continuous, Linear Model}

Recall that discretized, linearized dynamics are required in order to propagate the state estimate covariance and perform
the update step. It is our preference to linearize and then discretize; this is the approach found in many texts. Linearization is performed by expanding each state around its current estimate using a first-order approximation.\footnote{For the full derivation of the linearized system presented in this section, see \url{http://www-clmc.usc.edu/~nrotella/IROS2014_linearization.pdf}} This approach results in the linearized model
\begin{align}
\dot{\delta r} &= \delta v\\
\dot{\delta v} &= -C^{T}f^{\times}\delta \phi - C^{T}\delta b_{f} - C^{T}w_{f}\\
\dot{\delta \phi} &= -\omega^{\times}\delta \phi - \delta b_{\omega} -
w_{\omega}\\
\dot{\delta p} &= C^{T}w_{p}\\
\dot{\delta b}_{f} &= w_{bf}\\
\dot{\delta b}_{\omega} &= w_{b\omega}\\
\dot{\delta \theta} &= w_{z}
\end{align}
where the measured IMU quantities are bias-compensated and where $v^{\times}$ denotes the skew-symmetric matrix corresponding to the vector $v$.  The error state and process noise vectors are defined to be $\delta x = [\delta r, \delta v, \delta \phi, \delta p, \delta b_{f}, \delta b_{\omega}, \delta \theta]^{T}$ and $w=[w_{f}, w_{\omega}, w_{p}, w_{bf}, w_{b\omega}, w_{z}]^{T}$, respectively.  The linearized system can then be written in state-space form as $\dot{\delta x} = F_{c}\delta x + L_{c}w$ where
\begin{equation*}
F_{c} =
\begin{pmatrix}
0 & I & 0 & 0 & 0 & 0 & 0\\
0 & 0 & -C^{T}f^{\times} & 0 & -C^{T} & 0 & 0\\
0 & 0 & -\omega^{\times} & 0 & 0 & -I & 0\\
0 & 0 & 0 & 0 & 0 & 0 & 0\\
0 & 0 & 0 & 0 & 0 & 0 & 0\\
0 & 0 & 0 & 0 & 0 & 0 & 0\\
0 & 0 & 0 & 0 & 0 & 0 & 0
\end{pmatrix}
\end{equation*}
and $L_{c} = diag\{-C^{T}, -I, C^{T}, I, I, I\}$ are the prediction and noise Jacobians, respectively.  It is assumed for simplicity that the covariance matrix of each
process noise vector is diagonal with equal entries.  The continuous process noise covariance matrix is then $Q_{c} = diag\{Q_{f}, Q_{\omega}, Q_{p}, Q_{b_{f}}, Q_{b_{\omega}}, Q_{z}\}$.
The measurement model defined by \eqref{eq:pos_meas} and \eqref{eq:rot_meas} is linearized as
\begin{align*}
s_{p} &= -C\delta r + (C(p - r))^{x}\delta
\phi + C\delta p + n_{p}\\
s_{z} &=  \delta \phi - C[q\otimes z^{-1}]\delta \theta + n_{z}
\end{align*}
where $C[m]$ is used to denote the rotation matrix corresponding to the
quaternion $m$.
This model can be written in the form $\delta y = H_{c}\delta x + v$ where $v = [n_{p},
n_{z}]^{T}$ is the measurement noise vector and
\begin{equation*}
H_{c} = 
\setlength{\arraycolsep}{2.5pt}
\begin{pmatrix}
-C & 0 & (C(p - r))^{x} & C & 0 & 0 & 0\\
0 & 0 & I & 0 & 0 & 0 & -C[q\otimes z^{-1}]
\end{pmatrix}
\end{equation*}
is the measurement Jacobian.  It is assumed that measurements
are uncorrelated and hence the measurement noise covariance matrix is defined as
$R_{c} = diag\{R_{p}, R_{z}\}$ where $R_{p}$ and $R_{z}$ are again each diagonal
with equal entries.

\subsection{Discrete, Linear Model}

Assuming a zero-order hold on inputs over the interval $\Delta t = t_{k+1}-t_{k}$, the
discretized prediction Jacobian is given by
\begin{equation*}
F_{k} = e^{F_{c}\Delta t}
\end{equation*}
and the discretized state covariance matrix is given by
\begin{equation*}
Q_{k-1} = \int_{t_{k-1}}^{t_{k}}
e^{F_{c}(t_{k}-\tau)}L_{c}Q_{c}L_{c}^{T}e^{F_{c}^{T}(t_{k}-\tau)}d\tau
\end{equation*}
In practice, these expressions are often truncated at first order for simplicity to yield $F_{k} \approx
I + F_{c}\Delta t$ and $Q_{k} \approx F_{k}L_{c}Q_{c}L_{c}^{T}F_{k}^{T}\Delta t$.  In contrast to \cite{main}, the system is discretized using a zero-order hold on the noise terms as above in order to simplify the implementation. 

Following the above procedure, we find the discretized prediction and measurement Jacobians to be
\begin{equation*}
F_{k} = 
\begin{pmatrix}
I & I\Delta t & 0 & 0 & 0 & 0 & 0\\
0 & I & -C_{k}^{T}f_{k}^{\times}\Delta t & 0 & -C_{k}^{T}\Delta t & 0 & 0\\
0 & 0 & I-\omega_{k}^{\times}\Delta t & 0 & 0 & -I\Delta t & 0\\
0 & 0 & 0 & I & 0 & 0 & 0\\
0 & 0 & 0 & 0 & I & 0 & 0\\
0 & 0 & 0 & 0 & 0 & I & 0\\
0 & 0 & 0 & 0 & 0 & 0 & I
\end{pmatrix},
\end{equation*}
\begin{equation*}
H_{k} = 
\setlength{\arraycolsep}{2.5pt}
\begin{pmatrix}
-C_{k} & 0 & (C_{k}(p_{k} - r_{k}))^{\times} & C_{k} & 0 & 0 & 0\\
0 & 0 & I & 0 & 0 & 0 & -C[q_{k}\otimes z_{k}^{-1}]
\end{pmatrix}
\end{equation*}
where all quantities are computed using the \emph{a priori} state vector $\hat{x}_{k}^{-}$.
Finally, the continuous measurement covariance matrix is discretized as $R_{k-1}
\approx \frac{R_{c}}{\Delta t}$.

\subsection{Observability Analysis}

\begin{figure*}
\centering
\begin{subfigure}[b]{\textwidth}
\begin{tikzpicture}[spy using outlines={rectangle,yellow,magnification=4,width=1.5cm,height=1.5cm,connect spies}]
\node {\includegraphics*[height=.26\textheight, width=.99\textwidth, clip=true, trim=140 40 150 37]{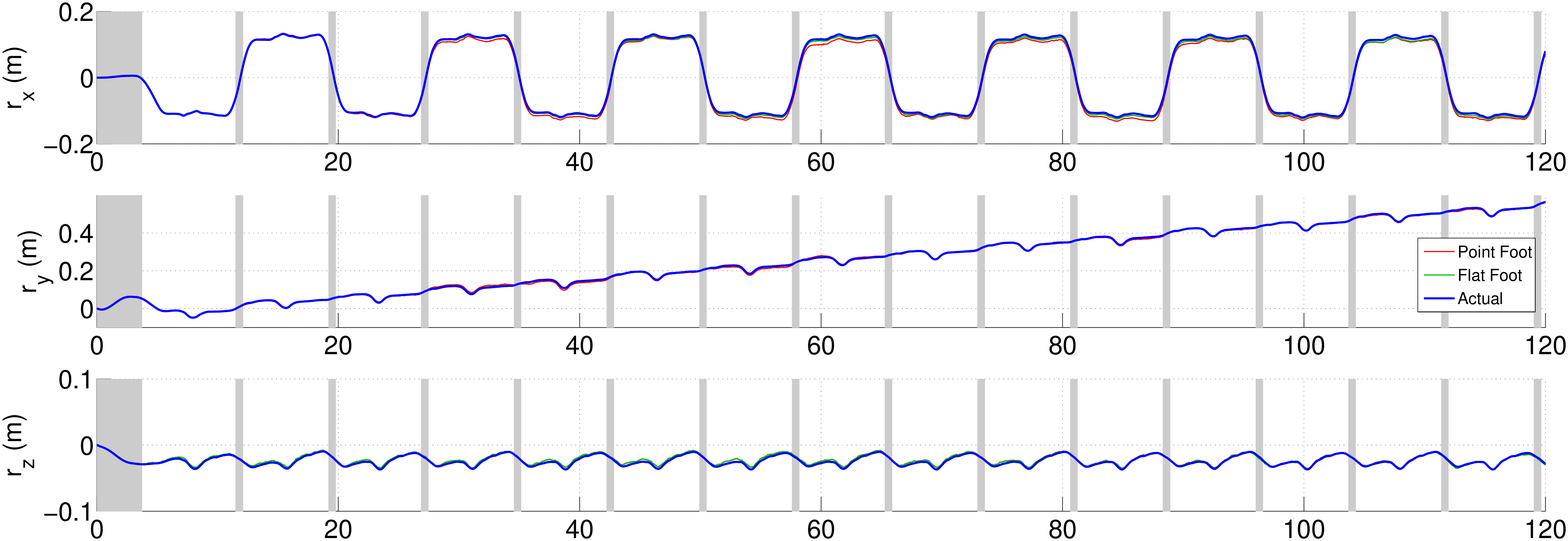}};
\spy on (0.5,2.7) in node at (4.4,-1);
\spy on (-2.5,-0.15) in node at (-3.9,0.9);
\end{tikzpicture}
\caption{Position}
\end{subfigure}
\begin{subfigure}[b]{\textwidth}
\begin{tikzpicture}[spy using outlines={rectangle,yellow,magnification=4,width=1.5cm,height=1.5cm,connect spies}]
\node {\includegraphics*[height=.27\textheight, width=.99\textwidth, clip=true, trim=140 40 150 38.5]{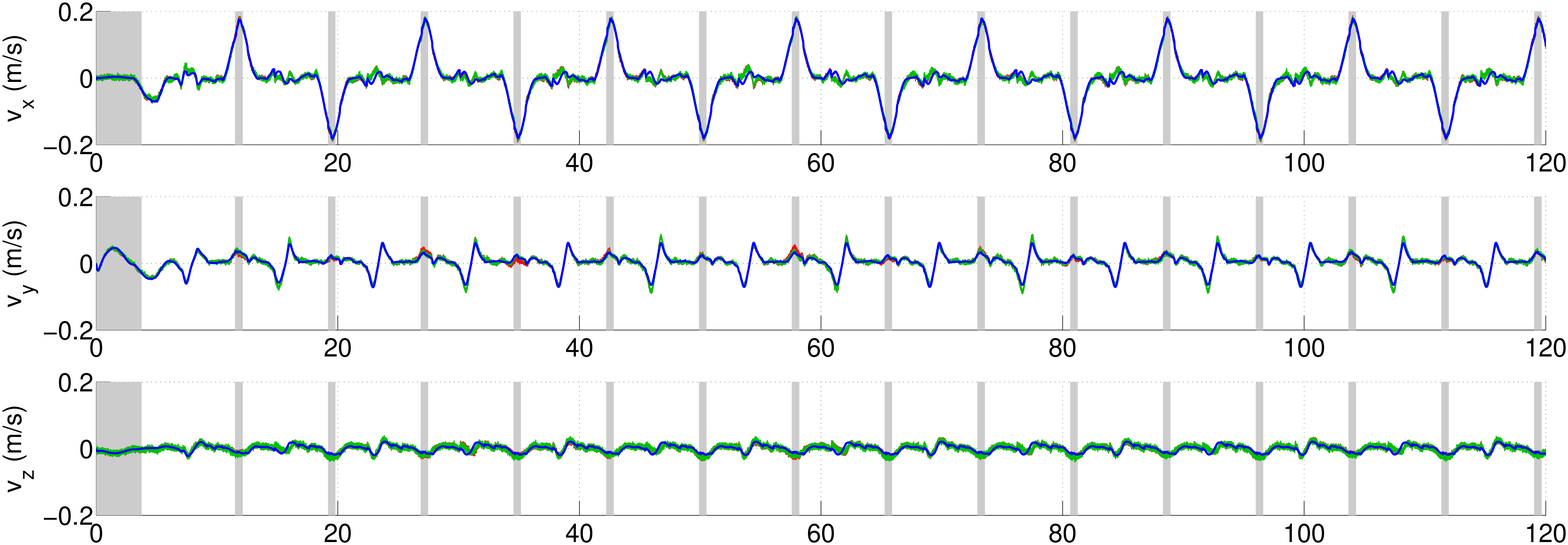}};
\spy on (-2.45,2.35) in node at (2.2,1.2);
\spy on (-4,.2) in node at (-6.5,-1);
\end{tikzpicture}
\caption{Velocity}
\end{subfigure}
\begin{subfigure}[b]{\textwidth}
\begin{tikzpicture}[spy using outlines={rectangle,yellow,magnification=4,width=1.5cm,height=1.5cm,connect spies}]
\node {\includegraphics*[height=.30\textheight, width=.99\textwidth, clip=true, trim=140 40 150 37]{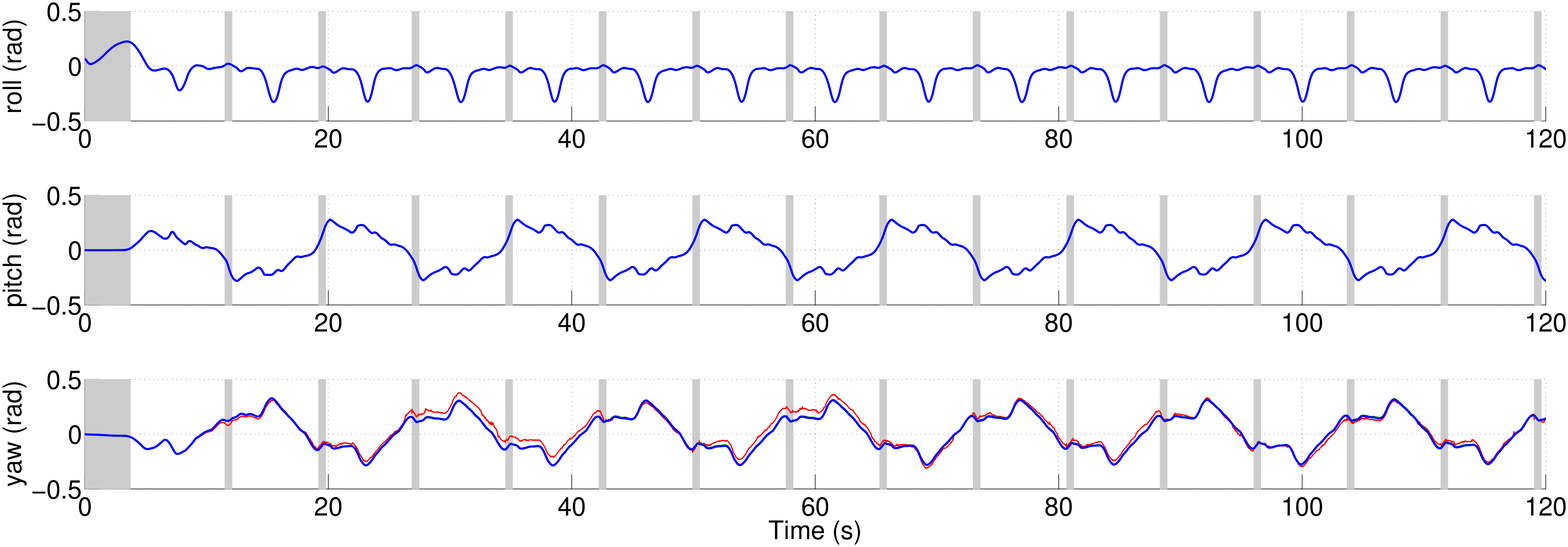}};
\spy on (-3.6,-1.82) in node at (1.4,-0.9);
\end{tikzpicture}
\caption{Euler Angles}
\end{subfigure}
\caption{Position, velocity and orientation (Euler angle) estimates for the 120 second walking task.  The portions with a white background indicate the single support phase while those with a grey background indicate the double support phase.  The yellow boxes highlight some periods of divergence discussed in the results section.  Overall, the flat foot filter introduced in this work performs better than the filter introduced in \cite{main} as confirmed by the estimation errors listed in Table \ref{tab:Errors}.}
\label{fig:walk}
\end{figure*}

As in \cite{main}, a nonlinear observability analysis of the filter was performed.  This section
discusses the resulting observability characteristics and analyzes the theoretical advantages of the addition of the foot
rotation constraints.  

The unobservable subspace informally describes all directions along
which state disturbances cannot be observed in the outputs. This corresponds to the nullspace of the observability matrix. For a nonlinear system, determining this matrix involves computing the gradient of successive Lie derivatives of the measurement model \eqref{eq:nonlin_meas} with
respect to the process model \eqref{eq:nonlin_pred} as in \cite{hermann}.  Since this space depends on the robot's motion, the presented analysis reveals all possible singularities and corresponding rank losses (RL). Note that since absolute position and yaw are inherently unobservable in this system, rank loss here represents an increase in the dimension of the unobservable subspace
beyond this nominal case.   While the full derivation\footnote{For the full derivation of the observability analysis presented in this section, see \url{http://www-clmc.usc.edu/~nrotella/IROS2014_observability.pdf}} is omitted for brevity, the pertinent results are summarized in tables
\ref{tab:pointFootObsv}, \ref{tab:twoPointFootObsv}, and \ref{tab:flatFootObsv}.

Table \ref{tab:pointFootObsv} below describes the rank deficiency for the case
in which a single point foot is in contact. The top row ($w = 0$) describes the case in which there is no rotational motion. Depending on the acceleration, the rank loss is either 3 or 5. The second row
($w \perp Cg$) states that rotational motion around an axis
which is perpendicular to the gravity axis leads to a rank loss of 1. The
third row ($w \parallel Cg$) describes the case in
which there is rotation only around the gravity axis; in general, this
corresponds to a rank loss of 1.
If the axis of rotation additionally intersects the point of
contact, rank loss increases to 2.  Further, if the IMU is directly above the
point of contact then rank loss increases to 3.
Finally, the last row summarizes the nominal case.

\begingroup{
\setlength{\tabcolsep}{.16667em}
\begin{table}[h]
	\centering
	\begin{tabular}{|c|c|c|l|}
		\hline
		Rotation & Acceleration/Velocity & Foothold & RL \\
		\hline
		\multirow{2}{*}{$w = 0$} & $a = -1/2 g$ & $*$ & 5 \\
		\cline{2-4}
		 & $a \neq -1/2 g$ & $*$ & 3 \\
		\hline
		$w \perp C g $ & $*$ & $*$ & 1 \\
		\hline
		\multirow{3}{*}{$w \parallel C g $} & \multirow{2}{*}{$ \wedge \begin{array}{l} a = \left(C^T w\right) \times v \\ v = \left(C^T w\right) \times (r-p) \end{array}$} & $(r-p) \parallel g$ & 3 \\
		\cline{3-4}
		 & & $(r-p) \not\parallel g$ & 2 \\
		\cline{2-4}
		 & $ \vee \ \begin{array}{l} a \neq \left(C^T w\right) \times v \\ v \neq \left(C^T w\right) \times (r-p) \end{array}$ & $*$ & 1 \\
		\hline
		$\wedge \begin{array}{l} w \not\perp C g \\ w \not\parallel C g \end{array}$ & $*$ & $*$ & 0 \\
		\hline
	\end{tabular}
	\caption{Rank deficiency for a single point foot contact.}
	\label{tab:pointFootObsv}
\end{table}
}

Table \ref{tab:twoPointFootObsv} below details the singular cases for two point foot
contacts.  These are similar to the single point foot contact cases
but with reduced rank losses.

\begin{table}[h]
	\centering
	\begin{tabular}{|c|c|l|}
		\hline
		Rotation & Footholds & RL \\
		\hline
		\multirow{2}{*}{$w = 0$} & $2a + g \parallel \Delta p$ & 3 \\
		\cline{2-3}
		 & $2a + g \not\parallel \Delta p$ & 2 \\
		\hline
		$w \perp C g $ & $*$ & 1 \\
		\hline
		\multirow{2}{*}{$w \parallel C g $} & $g \parallel \Delta p$ & 1 \\
		\cline{2-3}
		 & $g \not\parallel \Delta p$ & 0 \\
		\hline
		$\wedge \begin{array}{l} w \not\perp C g \\ w \not\parallel C g \end{array}$ & $*$ & 0 \\
		\hline
	\end{tabular}
	\caption{Rank deficiency for two point foot contacts.}
	\label{tab:twoPointFootObsv}
\end{table}

In comparison to the point foot cases, the flat foot case is significantly simpler as shown in Table \ref{tab:flatFootObsv}.  These results are valid for any number of flat foot contacts since a
single flat foot contact fully constrains the pose of the base.

\begin{table}[h]
	\centering
	\begin{tabular}{|c|l|}
		\hline
		Rotation & RL \\
		\hline
		$w = 0$ & 2 \\
		\hline
		$w \perp C g $ & 1 \\
		\hline
		$w \not\perp C g $ & 0 \\
		\hline
	\end{tabular}
	\caption{Rank deficiency for an arbitrary number of flat foot contacts.}
	\label{tab:flatFootObsv}
\end{table}

The rank loss of the new filter depends only on the base angular velocity.  If the angular velocity is zero then the rank loss is 2; if the axis of rotation is perpendicular to the gravity axis then the
rank loss is 1.  For all other cases, only the absolute
position and yaw are unobservable.

It's clear that the additional information resulting from the the rotational
constraint of the foot significantly reduces rank loss, as expected. In summary,
 the maximum rank loss is reduced from 5 to 2 when there is no rotational
 motion.  Additionally, rotation purely around the gravity axis no longer
 induces rank loss.  Finally, since the results of Table \ref{tab:flatFootObsv} hold for any number of flat foot contacts, both single and double support walking phases have desirable observability characteristics.  The experimental results of section \ref{sec:results} demonstrate the practical effects of the singular cases.

\section{Experimental Setup}

The goal of this work is to implement the proposed EKF on a SARCOS
humanoid equipped with a Microstrain 3DM-GX3-25 IMU.  However, it was desired that its performance first be verified in simulation.  For this purpose, both filters were implemented in the SL 
simulation environment \cite{schaal}.

In order to provide an accurate assessment of these filters, realistic levels of noise were added in the simulator by drawing samples from an i.i.d. Gaussian white noise process. Thermal noise was added to the
simulated IMU sensor data along with an integrated random walk bias.  Similarly, measurement noise
was added to the measurements at each
timestep. The standard deviations of the noise processes are given in Table \ref{tab:noiseParams}; all but the measurement densities are given in continuous time and are converted to discrete variances by squaring them and dividing by the timestep.

\begin{table}[h]
	\centering
	\begin{tabular}{|c|c|c|c|}
		\hline
		$w_{f}$ & $0.00078 m/s^{2}/\sqrt{Hz}$ & $w_{p}$ &  $0.001 m/\sqrt{Hz}$ \\
		\hline
		$w_{\omega}$ & $0.000523 rad/s/\sqrt{Hz}$ & $w_{z}$ &  $0.01 rad/\sqrt{Hz}$ \\
		\hline
		$w_{bf}$ & $0.0001 m/s^{3}/\sqrt{Hz}$ & $n_{p}$ & 0.01 m\\
		\hline
		$w_{b\omega}$ & $0.000618 rad/s^{2}/\sqrt{Hz}$ & $n_{z}$ & 0.01 rad\\
		\hline
	\end{tabular}
	\caption{Simulated noise parameters.}
	\label{tab:noiseParams}
\end{table}

The simulated IMU noise parameters were derived directly from the 3DM-GX3-25 datasheet and the measurement noise parameters were based on the observed uncertainty in the encoders and kinematic model of the
actual robot. Experiments were conducted at an update rate of 1000Hz as this is the fastest possible IMU streaming rate.

The measurement noise parameters were empirically tuned from their simulated values; all other parameters were set to their simulated values.  Initialization of the base orientation was performed using stationary accelerometer measurements, initial foot poses were computed from kinematics and all other states were initialized to zero.
 
\section{Results and Analysis}\label{sec:results} 
 
The plots in Figure \ref{fig:walk} show the results obtained on the simulated walking dataset of length
120 seconds.  Table \ref{tab:Errors} below lists the RMS and maximum errors for all plotted quantities with the last three rows corresponding to the Euler angles roll, pitch and yaw.  

\renewcommand{\arraystretch}{1}%
\begin{table}[h]
	\centering
	\begin{tabular}{|c|c|c|c|c|}
		\hline
		 & \multicolumn{2}{|c|}{RMS Error} & \multicolumn{2}{|c|}{Max Error}\\
		 \hline
	      & Point & Flat & Point & Flat \\
	     \hline
		 $r_{x} (m)$ & $0.0088$ & $0.0042$ & $0.0172$ & $0.0086$\\
	     \hline
	     $r_{y} (m)$ & $0.0046$ & $0.0017$ & $0.0123$ & $0.0047$\\
	     \hline
	     $r_{z} (m)$ & $0.0020$ & $0.0019$ & $0.0051$ & $0.0050$\\
	     \hline
	     $v_{x} (m/s)$ & $0.0079$ & $0.0082$ & $0.0417$ & $0.0393$\\
	     \hline
	     $v_{y} (m/s)$ & $0.0058$ & $0.0053$ & $0.0282$ & $0.0276$\\
	     \hline
	     $v_{z} (m/s)$ & $0.0067$ & $0.0066$ & $0.0357$ & $0.0321$\\
	     \hline
	     $\alpha (rad)$ & $0.0011$ & $0.0011$ & $0.0037$ & $0.0038$\\
	     \hline
		 $\beta (rad)$ & $0.0010$ & $0.0013$ & $0.0034$ & $0.0046$\\
		 \hline
		 $\gamma (rad)$ & $0.0379$ & $0.0055$ & $0.1032$ & $0.0110$\\
	     \hline
	\end{tabular}
	\caption{RMS and maximum (absolute) error values for the 120 second walking task for point and flat foot filters.}
	\label{tab:Errors}
\end{table}

Based on the above errors, the two filters perform equally well for many of the plotted quantities.  However, investigation of the velocity estimation reveals that the point foot filter periodically diverges more drastically throughout the task, causing the flat foot position estimates to be noticeably more accurate as demonstrated by the plots and error values.  Indeed, the maximum absolute errors in $v_{x}$, $v_{y}$ and $v_{z}$ for the point foot filter are reduced for the flat foot filter as compared to the point foot filter as shown in Table \ref{tab:Errors}.  While small, these repeated periods of divergence can lead to considerably more integrated error over the course of a lengthy task.

The divergence of the point foot filter in $v_{y}$ corresponds primarily to the double support portions of the task.  There is relatively little base rotation during these intervals since the robot is shifting its center of mass in preparation for the next step; proximity to the $\omega=0$ singular case may account for the performance difference between the filters since the flat foot filter has a reduced maximum rank loss in this situation. Both filters diverge in $v_{x}$ during the single support phase, suggesting that one or more singular cases are reached here (for example, the angular velocity becomes nearly orthogonal to the gravity axis when the leg is swung forward).  However, it's clear from the plots that the point foot filter provides poorer state estimates during this interval; this is to be expected since the rank loss cases for the flat foot filter are fewer and less drastic during single support.  Additionally, the large difference in error values between filters for the yaw angle is explained by the fact that constraining the rotational state renders the gyroscope bias fully observable for the flat foot filter (see the detailed observability analysis). Finally, note that the setup simulated in this paper employs high-quality sensors and a fast update rate; the difference between the two formulations is expected to be even more pronounced on a robot which has a lower control frequency, a less-accurate kinematic model or low-grade sensors.  The effect of hardware on performance remains to be tested in future work by adjusting simulated noise appropriately.

\section{CONCLUSIONS}

This paper introduces an EKF for state estimation on humanoid robots which builds on the filter introduced in \cite{main}.  These extensions to the previous work are shown to improve the observability characteristics of the system as well as improve performance on a realistic simulated platform.  In future work, the filter will be implemented and verified on the robot through extensive testing in balancing and locomotion tasks.  Additional extensions to the presented framework involving new sensors will be investigated and the resulting formulations will be compared against the presented filter.

\addtolength{\textheight}{-12cm}   


\bibliographystyle{IEEEtran} 
\bibliography{mybibfile}

\end{document}